\documentclass[letterpaper]{article} 
\usepackage{aaai2026}  
\usepackage{times}  
\usepackage{helvet}  
\usepackage{courier}  
\usepackage[hyphens]{url}  
\usepackage{graphicx} 
\usepackage{natbib}  
\usepackage{caption} 
\usepackage{footmisc}
\usepackage{color}
\usepackage{algorithm}
\usepackage{algorithmic}

\usepackage{subcaption}
\usepackage{multirow}
\usepackage{amsmath}
\usepackage{amsfonts}

\usepackage{newfloat}
\usepackage{listings}
\DeclareCaptionStyle{ruled}{labelfont=normalfont,labelsep=colon,strut=off} 
\floatstyle{ruled}
\newfloat{listing}{tb}{lst}{}
\floatname{listing}{Listing}
\title{Beyond Patches: Mining Interpretable Part-Prototypes for Explainable AI}
\author{
    Mahdi Alehdaghi\textsuperscript{\rm 1},
    Rajarshi Bhattacharya\textsuperscript{\rm 1},
    Pourya Shamsolmoali\textsuperscript{\rm 2}, \\
    Rafael M. O. Cruz\textsuperscript{\rm 1},
    Maguelonne Heritier\textsuperscript{\rm 3},
    Eric Granger\textsuperscript{\rm 1}
}
\affiliations{
    \textsuperscript{\rm 1}LIVIA, Dept. of Systems Engineering, ETS Montreal, Canada\\

    \textsuperscript{\rm 2}Dept. of Computer Science, University of York, UK \\
    \textsuperscript{\rm 3}Genetec Inc. 

    \{mahdi.alehdaghi.1, rajarshi.bhattacharya.1\}@ens.etsmtl.ca,  pshams55@gmail.com, mheritier@genetec.com, \\ \{rafael.menelau-cruz, eric.granger\}@etsmtl.ca
}

\usepackage{bibentry}

\begin{document}

\maketitle

\begin{abstract}
%
%

As AI systems grow more capable, it becomes increasingly important that their decisions remain understandable and aligned with human expectations. A key challenge is the limited interpretability of deep models. Post-hoc methods like GradCAM offer heatmaps but provide limited conceptual insight, while prototype-based approaches offer example-based explanations but often rely on rigid region selection and lack semantic consistency. 
To address these limitations, we propose PCMNet, a part-prototypical concept mining network that learns human-comprehensible prototypes from meaningful image regions without additional supervision. By clustering these prototypes into concept groups and extracting concept activation vectors, PCMNet provides structured, concept-level explanations and enhances robustness to  occlusion and challenging conditions, which are both critical for building reliable and aligned AI systems. 
Experiments across multiple image classification benchmarks show that PCMNet outperforms state-of-the-art methods in interpretability, stability, and robustness. This work contributes to AI alignment by enhancing transparency, controllability, and trustworthiness in AI systems. 
Our code is available at: \url{https://github.com/alehdaghi/PCMNet}.
\end{abstract}

\section{Introduction}
\label{sec:intro}

Deep learning (DL) has advanced computer vision, enabling models to achieve remarkable performance across a wide range of tasks, including object detection and image classification \cite{chai2021deep}. However, despite these successes, DL models often operate as black boxes, and their lack of transparency has raised critical concerns regarding interpretability and reliability. This problem becomes even more significant in areas such as healthcare \cite{fellous2019explainable}, self-driving cars \cite{kim2022xai}, and security systems \cite{ProtoPnet, PIPNet}, where accurate predictions are not sufficient; it is crucial to understand the reasoning behind them. In these situations, explaining what the model is doing is important. It helps people trust the system, find problems if something goes wrong, and make sure the system follows important rules and laws \cite{markus2021role}. Therefore, researchers are now attempting to develop DL models that not only perform well but also provide interpretable and human-understandable explanations for their decisions \cite{saranya2023systematic, vilone2020explainable}.

\begin{figure*}
    \centering
    \centering
  \includegraphics[trim={0 0.0cm 0 0.05cm},clip,width=\textwidth]{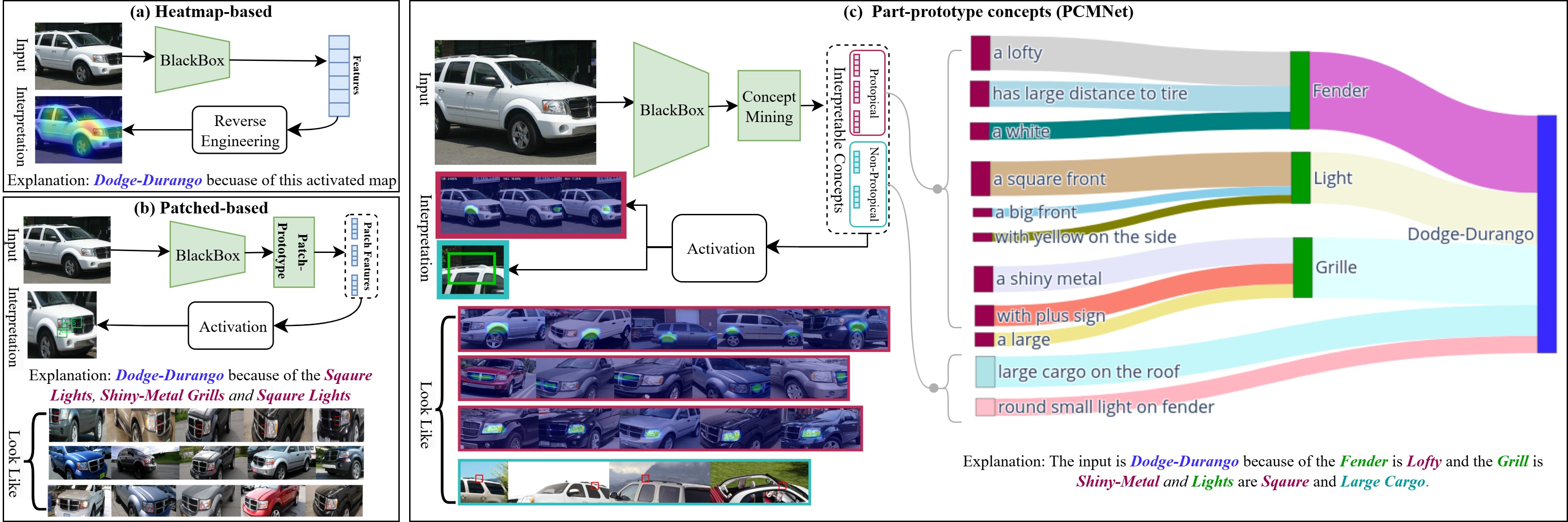}
\vspace{-0.65cm}
\caption{Explanation methods. (a) Heatmap-based methods see the feature backbones as black-box models, and try to locate regions activated by the most important features. (b) Patch-based methods decompose the input into image patches and explain the decision using patch-derived components. (c) PCMNet, in contrast, mines both prototypical and non-prototypical concepts, offering a broader and more interpretable set of regions to explain the model decision.}
\label{fig:teaser}
\end{figure*}

Common post-hoc explainability methods, such as GradCAM \cite{chakraborty2022generalizing} and ScoreCAM \cite{wang2020score}, aim to explain a model’s decision by identifying which parts of the input image had the most influence on the output. These methods analyze the trained model's gradients or activations to identify salient regions of the input. An example is shown in Fig. \ref{fig:teaser} (a).
Unlike these reverse-engineering methods,  ante-hoc methods are designed to make the model explainable from the beginning. They guide the model to produce either global \cite{he2025v2c,pcx, jiang2025knowledge} or local \cite{PIPNet,ayoobi2025protoargnet,ProtoTree, zhu2024enhancing} interpretable parts that help explain its decisions. This idea is inspired by the recognition-by-components theory \cite{biederman1987recognition}, which suggests that humans understand objects by breaking them down into meaningful parts or concepts. 

Some methods, such as ProtoPNet \cite{ProtoPnet} and PIPNet \cite{PIPNet}, as illustrated in Fig. \ref{fig:teaser} (b), divide the spatial deep features (prior to the final pooling layer) into fixed, small patches and learn a set of prototypes from these regions. These prototypes, activated by specific input patches, serve as interpretable visual features that support the model’s decision. This also enables users to compare similar semantic concepts across different images, enhancing transparency and trust in the model's reasoning capacity.
However, these methods have notable limitations. Fixed patch sizes result in unstable prototype activation, particularly for larger semantic regions, where the patches fail to capture coherent or meaningful visual concepts. Conversely, very small patches lack sufficient contextual information, reducing their ability to represent interpretable or semantically rich components. Furthermore, discriminative and interpretable visual concepts often require larger regions to be represented clearly. When the model uses only small, fixed patches, it may fail to capture the full meaning of these concepts. This can lead to repetitive or shallow explanations, as seen in the bottom two rows of Fig. \ref{fig:teaser} (b), where similar patterns appear multiple times. To address this, the regions used for prototypes should not be fixed. Instead, they should be adjustable and learned during training, allowing the model to focus on more meaningful and diverse parts of the image.

One possible direction is to discover meaningful regions using unsupervised part-based or slot-based mechanisms. However, direct application of naïve part-discovery methods \cite{van2023pdisconet} or Slot Attention techniques \cite{monet2021object, locatello2020slot} fails to produce semantically meaningful and interpretable components. These approaches typically aim to group similar visual parts across instances, but do not enforce conceptual consistency, resulting in ambiguous or fragmented representations \cite{ProtoShare}. Moreover, Slot Attention-based features often lack explicit prototypical structures, making it difficult to associate them with clear and intuitive explanations \cite{li2021scouter}. As a result, the learned prototypes tend to lack diversity and interpretability, limiting the model ability to generate meaningful justifications for its predictions.

To address these limitations, we propose the Part-Prototypical Concept Mining Network (PCMNet), an interpretable image classifier that explains its predictions by extracting meaningful, part-based prototypical concepts \footnote{We define \textbf{part-prototypes} as localized visual regions that commonly appear across inputs (e.g., a car’s light), while the term \textbf{concept} describes a specific attribute or variation of that part (e.g., \emph{rectangular} or \emph{circular} shapes of the light).} from selected regions. PCMNet first identifies relationships between semantically related prototypes across different instances, even across class boundaries and then derives a set of primitive class-discriminative concepts to form the basis for the model decision-making process.

In the first stage, PCMNet uses an unsupervised part-discovery module that learns prototypes through a novel center-clustering loss. In the second stage, deep features are encoded into a sparse Concept Activation Vector (CAV)~\cite{CAV}, which provides a spatially and semantically interpretable representation of the model’s decision. Each concept activation is computed as the cosine similarity between part features and class-aware prototype centroids. To ensure both conceptual simplicity and class-specific relevance, PCMNet applies a clustering algorithm to the part features associated with each prototype, and abstracts each cluster into a representative concept vector. The Concept Activation Vector (CAV) produced by PCMNet is inherently interpretable, as it is generated through a sparse, non-negative linear transformation that ensures all concept scores are positive. As illustrated in Fig.~\ref{fig:teaser} (c), PCMNet activates concepts such as “lofty fenders,” “shiny metal grille,” and “square-shaped lights” to support its prediction that the vehicle is a Dodge Durango, offering a clear and meaningful explanation of the model’s reasoning.



The contributions of this paper are summarized as follows.
\textbf{(1)} We propose a part-prototypical concept mining network (PCMNet) that jointly learns adaptive semantic regions and class-discriminative prototypes for interpretable image classification. 
\textbf{(2)} A contrastive prototype learning mechanism is introduced that extracts semantically coherent concepts from adaptively sized image regions rather than fixed patches. 
\textbf{(3)} We develop a two-level clustering approach with pixel-level part discovery and feature-level concept formation. 
\textbf{(4)} PCMNet outperforms ProtoPNet and PIPNet in both occlusion robustness (+7.2$\%$ accuracy at 30$\%$ occlusion) and explainability across multiple datasets.

\section{Related Work}
\label{sec:related}
\vspace{-0.1cm}

\noindent \textbf{(a) Post-hoc or heatmap-based explanation. }
Heatmap-based explainability methods are a widely used family of post-hoc techniques that visualize which parts of an input image contribute most to a model’s decision. These techniques, often referred to as attribution methods, assign importance scores to different image regions to highlight influential features. Gradient-based methods, such as GradCAM~\cite{selvaraju2017grad} and ScoreCAM~\cite{wang2020score}, generate heatmaps by backpropagating gradients concerning input features, revealing class-specific activation regions. While GradCAM produces class-dependent heatmaps, other methods like FullGrad~\cite{srinivas2019fullgrad} are class-agnostic, providing a more general view of model behavior across different outputs. Despite their popularity, gradient-based methods suffer from high sensitivity to noise, which can lead to unreliable and inconsistent heatmaps. To mitigate this, gradient-free CAM techniques, such as ScoreCAM, have been proposed to generate more stable and interpretable explanation maps.

Beyond gradient-based methods, attribution propagation approaches offer an alternative way to decompose model predictions into layer-wise relevance scores. Techniques such as Layer-wise Relevance Propagation (LRP)~\cite{bach2015lrp} or KD-FMV \cite{jiang2025knowledge} recursively distribute relevance through the network, providing a structured breakdown of model decisions. While initially designed for convolutional neural networks (CNNs), recent adaptations have extended these methods to vision transformers (ViTs)~\cite{chefer2021transformer}, using their self-attention mechanisms for improved interpretability. However, despite their effectiveness in highlighting important image regions, both gradient-based and attribution propagation methods remain fundamentally limited in providing high-level, human interpretable concepts. Unlike our PCMNet, which explains decisions through part-based prototypes, heatmap-based methods do not inherently capture semantic relationships between features, making them less suitable for interpretable concept-based explanations.

\noindent \textbf{(b) Concept-based explanation.}
Concept-based XAI methods \cite{pcx,ProtoPnet,ProtoShare,bach2015lrp,he2025v2c} analyze the role of latent representations in specific layers of a deep neural network to explore concepts from input images and find Concept Activation Vectors (CAVs) \cite{CAV} for the made decision. Early XAI research primarily focused on understanding how these concepts contribute to global decision-making, identifying the most relevant concepts for a given output class \cite{pcx,bach2015lrp}. Some approaches enhance model interpretability by employing local feature attribution techniques to localize and quantify the importance of concepts for individual predictions, thereby enabling concept-based explanations at the instance level~\cite{PIPNet,ProtoTree,ayoobi2025protoargnet}. While these methods provide faithful explanations aligned with the model’s internal reasoning, they are vulnerable to performance degradation under occlusion, as their extracted concepts often rely on limited and highly localized discriminative regions of the input. PCMNet addresses this limitation by wider range of part-based prototypical concept reasoning to rely on other discriminative parts that are not occluded. 

Recently, Concept Bottleneck Models (CBMs) has emerged as a research avenue for inherently interpretable approaches that predict intermediate human-understandable concepts before class prediction \cite{xie2025discovering,posthocCBM,rao2024discover,he2025v2c}. Unlike patch- or part-prototype models, CBMs explicitly separate the reasoning process into concept prediction and concept-to-label mapping, allowing direct human intervention on the concept layer. However, their reliance on annotated concepts limits scalability, and the fidelity of the bottleneck depends on concept quality. Recent methods, including post-hoc CBMs \cite{he2025v2c,xie2025discovering} and vision-to-concept tokenizers that learn visual concept vocabularies without text supervision—aim to bridge this gap by enabling scalable and more interpretable reasoning.


\section{Prototypical Concept Mining Network}
\label{sec:method}

\begin{figure*}[!t]
    \centering
    \begin{subfigure}{0.75\linewidth}
         \includegraphics[trim={0.0cm 0.1cm 0.0cm 0.0cm},clip, width=\linewidth]{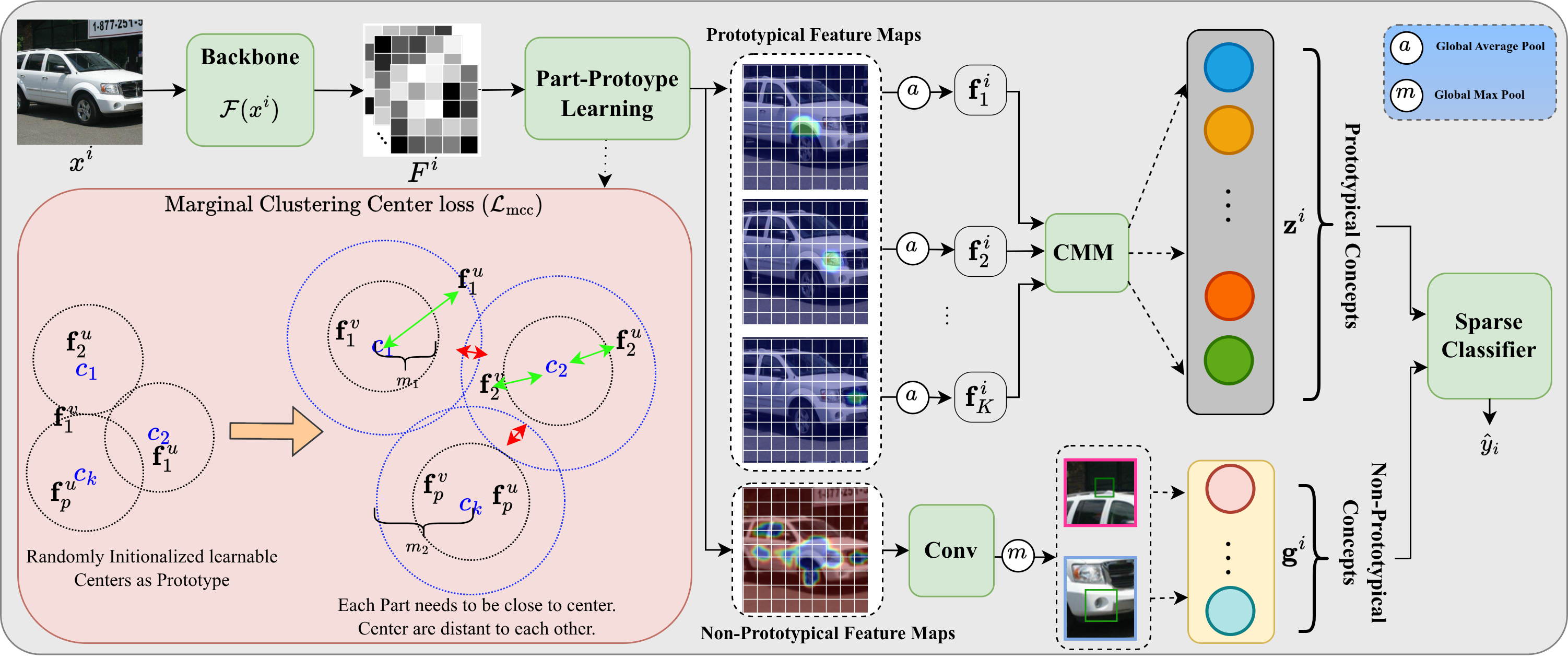} 
         \caption{PCMNet}
    \end{subfigure}
    \begin{subfigure}{0.22\linewidth}
         \includegraphics[trim={0.45 0.0cm 0 0.1cm},clip, width=\linewidth]{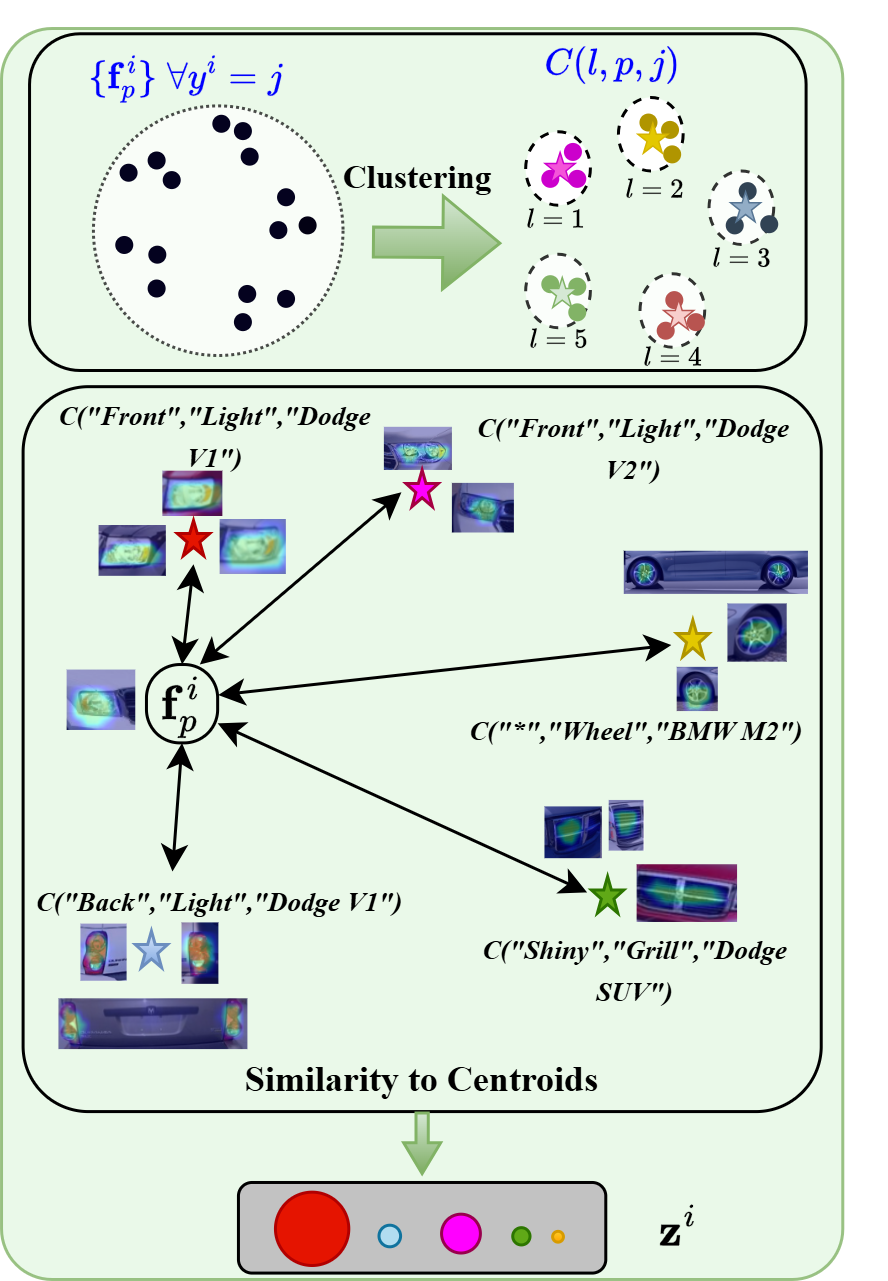}  
         \caption{CMM}
         \label{fig:cmm}
    \end{subfigure}
   
    \vspace{-0.3cm}
    
    \caption{(a) Overall architecture of PCMNet. In the first stage, part prototypes are learned from spatial features extracted by the backbone. In the second stage, prototypes are clustered within each class to identify frequently occurring concepts. The center of these clusters is computed as concept prototypes, and the distance between part prototypes and concept clusters determines the activated concepts for the model’s decision-making. This approach enables interpretable and concept-driven classification. (b) Concept Mining Module. First, a part centroid list is generated by applying DBSCAN\cite{DBSCAN} clustering within each class and computing the centers of the resulting clusters. These centroids serve as frequently occurring concept prototypes. Next, to activate the most relevant concept for a given part feature,  the similarity (inverse distance) to the centroids is measured. These values are then used as CAVs to inform the model’s decision-making. }
    \label{fig:model}
\end{figure*}

To explain classification decisions over a training set $\mathcal{T} = {(x^1, y^1), ..., (x^N, y^N)} \subset \mathcal{X} \times \mathcal{Y}$ with $L$ classes, PCMNet extracts sparse and primitive concepts that are both activated and discriminative for object recognition. These concepts are captured through a compact set of prototypes that recur across images in the dataset (e.g., lights, fenders, and grilles for cars; heads and wings for birds). The model outputs a class-based Concept Activation Vector (CAV) $\mathbf{z}^i = [\mathbf{z}^i_1; \dots; \mathbf{z}^i_K]$, where $;$ denotes concatenation and each $\mathbf{z}_p$ represents the activated concepts from part $p$.

Fig. \ref{fig:model} shows the overall overview of PCMNet, which transforms features extracted from the backbone into explainable and interpretable concepts in three steps. \noindent \textbf{Step 1}: PCMNet identifies semantically meaningful and independent regions from input images in an unsupervised manner by minimizing the Marginal Cluster Center loss ($\mathcal{L}_\text{mcc}$). This process learns part-prototype features $\mathbf{f}^i_p$ that are informative for class discrimination. 
\textbf{Step 2}: Once part-level features are obtained, the Concept Mining Module (CMM) $\mathcal{H}$ aggregates these features across all instances of each class to discover a set of shared elementary visual patterns. Cluster centers derived from this process form the list $\mathcal{C}$, representing the core prototypical concepts. \textbf{Step 3}: For each input, a Concept Activation Vector (CAV) $\mathbf{z}^i$ is computed by measuring the similarity between its extracted part features and the concept list $\mathcal{C}$. These CAVs are used both for classification and for explaining the model’s decision. To preserve class-specific but non-shared discriminative cues, PCMNet also incorporates features outside the prototypical regions into the CAVs.

\subsection{Step 1: Part-Prototype Discovery}
Our model is shown in Fig. \ref{fig:model}. The input images are given to a backbone to extract the 3D deep features as $F^i \in \mathbb{R}^{D\times w \times h}$ where $D$ is the channel size and $w, h$ denote the width and height. Based on extracted deep features $F^i$, our model finds semantically independent meaningful regions that can be shared in different inputs. We name them ``part-prototype''. For determining regions, the model scores each pixel in $F^i$ to specify their belongingness to each part-prototype class. The mapping score for each pixel $(a)$ is represented by $M_p^i(a)$, we have $\sum_{p=1}^{K+1} M^i_p(a) = 1$. 
The spatial feature for each part, $F^i_p$, is computed by element-wise multiplication of $M^i_p$ to $F^i$. Then part features, $\mathbf{f}^i_p$ is computed as:
\begin{equation}
\vspace{-0.1cm}
    \mathbf{f}^i_p = \mathcal{GP}(\texttt{conv}(F^i_p)) \in \mathbb{R}^{d_f},
\end{equation}
where $\mathcal{GP}$ is the global average pooling and $d_f$ is the dimension of features.
The last index (K+1) is used to cover the background or non-part regions of the foreground \cite{BMDG}. We name these features non-prototypical concepts that are computed as:
 \vspace{-0.1cm}
\begin{equation}
    \vspace{-0.2cm}
    \mathbf{g}^i = \mathcal{MP}(\texttt{conv}(F^i_{K+1})) \in \mathbb{R}^{d_f},
\end{equation}
where $\mathcal{MP}$ is the max-pooling. We use the PDiscoNet \cite{van2023pdisconet} model for extracting parts features. Our part loss comprises their final part-discovery loss ($\mathcal{L}_\text{pd}$) without their orthogonal component and with our marginal cluster center loss ($\mathcal{L}_\text{mcc}$): 
\vspace{-0.15cm}
\begin{equation}
\vspace{-0.15cm}
    \mathcal{L}_\text{part} =  \mathcal{L}_\text{bl} + \alpha \mathcal{L}_\text{mcc}. 
\end{equation}
where $\alpha$ is a parameter regularizing the effect of $\mathcal{L}_\text{mcc}$. 
The goal of $\mathcal{L}_\text{mcc}$ is to ensure that each selected region is consistent across inputs and captures semantically similar content, while remaining class-discriminative and identity-aware.

\subsubsection{Marginal Cluster Center Loss}
To ensure part-prototypes are both distinct and semantically meaningful, we encourage separation in the feature space while guiding part-level features to align with their corresponding prototypes. To handle intra-part variation across classes, we apply a soft-margin loss that allows features from the same part (but different classes) to stay within a bounded distance of their prototype. This encourages the model to discover consistent, class-aware part structures while preserving inter-class separation between prototypes. The loss can be written as
\begin{equation} \label{eq:cen}
    \begin{aligned} 
        \mathcal{L}_\text{mcc} \! = \! \sum_{p=1}^K [ |\mathbf{f}^i_p-c_p|\! - \!m_1 ]_\star\!
         +\!\frac{1}{K}\!\sum_{q\neq p}^K[ m_2\! -\! |c_p-c_q|]_\star,
    \end{aligned}
\end{equation}
where $c_p$ is a learnable prototype 
and $[\cdot]_\star$ is $\max(\cdot,0)$. 

\subsection{Step 2: Part-Centroid List Generation}

{In \textbf{Step 1}, the backbone learns to extract meaningful part-level features that support the classification task. To discover compact and consistent class-specific concepts, we collect all part features from the training set. Specifically, for each class $j$ and part $p$, we construct
$
B^j_p = \{ \mathbf{f}_p^i \mid y^i = j \} \in \mathbb{R}^{N_j \times d_f}
$, where $N_j$ is the number of training samples from class $j$.
To identify semantically coherent patterns, we apply the clustering algorithm separately to each $B^j_p$, yielding $T^j_p \in \mathbb{N}^{N_j}$, which assigns cluster labels to the part features.
We then compute the centroid of each cluster $l$ (representing a visual concept) as:
\begin{equation}
\vspace{-0.1cm}
\mathcal{C}(l, p, j) = \text{mean}(\{ B^j_p[ T^j_p = l ] \}) \in \mathbb{R}^{d_f}.
\end{equation}
The total number of discovered concepts across all classes and parts is computed as $d_c = \sum_{j=1}^L \sum_{p=1}^K \max(T^j_p)$.
This process, shown in the top of Fig.~\ref{fig:cmm}, allows the model to build a compact and expressive vocabulary of class-specific, part-aware visual concepts.


\subsection{Step 3: Concept Activation Mining}
Once the centroid list is generated, each concept's activation value is defined by the similarity of extracted features to each centroid: 
\begin{equation}
    \mathbf{z}^i =  \{\mathcal{S}(\mathcal{C}(l,p,j),\mathbf{f}_p^i) \; \forall p \in \{1..K\} \; \text{and} \; j \in \{1..L\}\}, 
\end{equation}
where $\mathbf{z}^i \in \mathbb{R}^{d_c}$ ans $\mathcal{S}$ is cosine distance. 
We aim to explain each output class of our model using a small set of interpretable concepts. We, therefore, train a sparse classifier on top of $z^i$ to obtain the final classification scores $o^i = W_1^T \mathbf{z}^i+ W_2^T \mathbf{g}^i + b$ and the predicted class $\hat{y}^i = \arg \max(o^i)$. Here, $W_1 \in \mathbb{R}^{d_c \times L}$, $W_2 \in \mathbb{R}^{d_f \times L}$  and $b \in \mathbb{R}^{L}$ denote the classification weights and bias term, respectively. This sparse layer is trained with the following sparse classification loss \cite{wong2021leveraging}:
\begin{equation}
    \mathcal{L}_\text{cl} = \mathcal{L}_{ce}( W_1^T \mathbf{z}^i+ W_2^T \mathbf{g}^i + b, y^i) + \lambda \mathcal{R}(W_1),
\end{equation}
where $\mathcal{L}_\text{ce}$ is the cross-entropy loss, $y^i$ is the label, $\lambda$ is the sparsity regularization strength and $\mathcal{R}(W) = (1-\gamma)\frac{1}{2}\Vert W \Vert_F + \gamma \Vert W \Vert_{1}$, here $\Vert W \Vert_F$ is the Forbenius norm and $\Vert W \Vert_1$ denotes the element-wise matrix.

\subsection{Overall Training}
We jointly optimize the network in an end-to-end manner:
\begin{equation}
\label{eq:all_losses}
    \mathcal{L} = \mathcal{L}_\text{part} + \beta \mathcal{L}_\text{cl},
\end{equation}
where $\beta$ is hyperparameter. We set $\beta=0$ during initial training and $\beta=2$ after learning part-prototypical concepts.


\section{Experiments and Results}
\label{sec:exp}

\begin{table*}[!t]
\centering
\resizebox{\textwidth}{!}{
\begin{tabular}{|l|l||c|c|c|c|c|c|}
\hline
 & \textbf{Method} & \textbf{Consistency (Intra) $\uparrow$} & \textbf{Consistency (Inter) $\downarrow$} & \textbf{F(1)-F(5) $\uparrow$} & \textbf{Sp $\uparrow$} & \textbf{Stability} & \textbf{C Acc $\uparrow$} \\ \hline \hline
\multirow{6}{*}{\rotatebox[origin=c]{90}{Cars-196}} & Baseline & \textbf{83.45 $\pm$ 2.7} & 27.26 $\pm$ 28.10 & 1.54 - 6.61 & 22.74 & 65.4 & 84.1 \\ 
 & ProtoPNet & 45.70 $\pm$ 4.9 & 11.22$\pm$5.5 & 9.3-80.12 & 60.92 & 69.6 & 84.5 \\ 
 
& PiPNet & 42.40$\pm$17.5 & 17.4$\pm$2.4 & 9.43 - \textbf{93.15} & 63.19 & 60.8 & 86.46 \\ 
& ProtoViT & 38.08$\pm$8.1 & 18.91$\pm$1.8 & 11.74 - 75.66 & 51.23 & 57.3 & \textbf{91.84} \\ 
& MCPNet & 48.91$\pm$14.6 & 9.17$\pm$3.2 & 14.78 - 88.02 & 60.42 & 64.26 & 80.15 \\ \cline{2-8} 
 & PCMNet (Ours) & 57.84$\pm$3.1 & \textbf{2.97 $\pm$ 1.37} & \textbf{59.33} - 91.73 & 57.45 & \textbf{71.5} & 90.15 \\ \hline \hline
\multirow{6}{*}{\rotatebox[origin=c]{90}{CUB-200}} & Resnet50   & 57.38 $\pm$ 3.50 & 25.45 $\pm$ 23.82 & 2.43 - 5.73       & 24.14    & 60.3  &          81.2 \\ 
& ProtoPNet         & 51.47 $\pm$ 5.8    &  10.16$\pm$9.2                                   & 9.36-75.50               & 79.51    &  63.6    & 81.45            \\ 
& PiPNet          & 42.28$\pm$3.2 &    15.57$\pm$15.7                                  & 9.40 - 89.77 &  77.83    & 65.9   & 82.0           \\ 
& ProtoViT          & 51.76$\pm$3.8 &    13.60$\pm$12.5                                  & 14.22 - 58.98 &  54.93    & 56.4   & 85.3           \\ 
& MCPNet          & 52.94$\pm$5.9 &    8.11$\pm$5.0                                  & 24.60 - 67.30 &  50.33    & 58.4   & 80.1           \\ \cline{2-8}
& PCMNet (Ours)  & \textbf{57.37$\pm$3.4}  & \textbf{3.37 $\pm$ 4.2}                                   &                  \textbf{54.34 - 89.84} &  58.67  &  \textbf{68.5}  & 85.1            \\
\hline
 
\end{tabular}
}
\vspace{-0.16cm}
\caption{Performance of PCMNet according to xAI metrics on the Stanford Cars and Birds datasets.}
\label{tab:cars}
\end{table*}

\subsection{Implementation Details}
{ 
PCMNet is evaluated on two standard prototype-learning benchmarks: Stanford Cars \cite{scar} (196 car classes) and CUB-200-2011 \cite{cubs} (200 bird species). We use a pretrained ResNet50 as the backbone. Each training batch (32 images) is resized to 488×488 with random cropping and padding. Optimization is performed using Adam \cite{kingma2014adam}. Regularization parameters $\lambda$ and $\gamma$ follow \cite{oikarinen2023label}; we set $\alpha = 1.5$, $\beta = 2$, with margins $m_1 = 0.3$ and $m_2 = 1.5$ based on hyperparameter tuning analysis on the Appendix A. 

\noindent \textbf{End-to-End Multi-Stage Training}:
PCMNet modules (part discovery, concept clustering, and concept activation) are  trained end-to-end in a unified pipeline over 40 epochs:

{\noindent\textit{- Step 1}} (initial epochs): The part discovery module is optimized to localize semantically meaningful part features. This stage continues until convergence of $\mathcal{L}_\text{part}$.

{\noindent\textit{- Step 2 }(intermediate epochs)}: Concept clustering runs intermittently (every 5 epochs) to reduce computation while keeping centroids up to date with changing representations.

{\noindent \textit{- Step 3} (final epochs):} The Concept Activation Mining Module (CMM) is trained alongside the backbone, using the updated centroids from Stage 2 to align activated features with discriminative concept prototypes.
At inference, PCMNet behaves as a standard forward-pass classifier.

\subsection{Explainability Metrics}

To evaluate explainability, we use established metrics from \cite{pcx}, with minor adjustments. 

\noindent \textbf{(a) Faithfulness:}  
Faithfulness \cite{fait1,fait2} evaluates how much the activated concepts influence the model’s final prediction, serving as a key metric for assessing the quality of explanations. One widely used approach for measuring faithfulness is concept deletion, which involves removing selected concepts from the model's reasoning process and observing the resulting change in output confidence. Prior works, such as \cite{pcx}, typically assess this by removing only the single most important concept. However, this limited scope fails to capture the broader contribution of other activated concepts and may not reveal the full extent of their impact.
To provide a more comprehensive assessment, we extend this approach by successively removing the top-$k$ most activated concepts and measuring the resulting drop in model confidence. A larger drop indicates higher faithfulness, as it suggests that the removed concepts were indeed critical to the model’s decision. This extended evaluation allows us to better quantify how much the explanation aligns with the model’s internal reasoning.

\noindent \textbf{(b) Stability:} 
The extracted concepts from input images that share some semantic concepts must be stable and explainable for unseen images. To evaluate stability, we compute CAVs on $k$-fold subsets of the data ($k$ = 10 as default) similar to \cite{pcx} for all classes at steps 2 and 3. We then map CAVs together using a Hungarian loss function and measure the cosine similarity between them.

\noindent \textbf{(c) Consistency:} 
Activated concepts from images with the same class should be similar to be consistent and explain the model's decision. To measure this consistency, we extract the CAV from images and then measure the cosine similarity between them. The mean cosine similarity is assessed between images from the same class and between images from different classes. The ratio between inter/intra-class similarity should be high for a stable explanation.

\noindent \textbf{(d) Sparseness}: 
The sparseness metric essentially describes the uniformity of the concept activation, where having some concepts activate more than others is considered easier to interpret. This is because a uniform distribution of concept activations would provide little information on the importance of specific concepts or parts of the images, i.e., a high entropy in the generated concepts.

\subsection{Comparison to Prototype xAI Methods}
To evaluate PCMNet, we report classification accuracy and XAI metrics, and compare them with other ante-hoc methods (ProtoPNet\cite{ProtoPnet}, PIPNet\cite{PIPNet}, ProtoViT \cite{protoViT} and MCPNet \cite{Wang2024MCPNet}) in Table \ref{tab:cars}. 
The consistency metric is reported in two forms: Intra and Inter, measuring the cosine similarity between activated concepts within the same class and across different classes, respectively.
Faithfulness is reported as \textbf{F(\textit{n})}, measuring accuracy drops when the top \textbf{\textit{n}} important concepts are removed from the decision.
To compute the contribution of each concept to the final decision, we compute a weighted score as the product of its raw value and its corresponding classification weight.

\subsection{Ablation Studies}

\subsubsection{(a) Effect of Number of Concept Center centroids ($d_c$): }
The size of the concept list ($d_c$) determines the semantic resolution of PCMNet and is influenced by both the number of parts and the number of training classes. As the number of parts ($K$) increases, so does the total number of extracted centroids. We analyze this dependency further in Appendix B. While richer concept sets can improve interpretability, they also increase memory and computational cost.

To manage this, we apply a post-clustering refinement step that merges similar centroids within or across classes using hierarchical clustering \cite{murtagh2011ward}. Table \ref{tab:ward_cluster} reports the effect of varying the merging level and threshold—defined as a percentage of the maximum inter-centroid distance—on the Cars dataset.
We find that moderate merging (e.g., Level 1 with 10\% threshold) significantly reduces $d_c$ with minimal impact on classification accuracy or faithfulness (F(3)). However, excessive merging (e.g., Level 3 at 10\%) leads to performance degradation, suggesting the importance of controlled prototype compression.

\begin{table}[h]
\centering
\resizebox{\linewidth}{!}{
\begin{tabular}{|c|c||c|c|c|}
\hline
\textbf{Threshold (\%)} & \textbf{Level} & \boldmath$d_c$ & \textbf{C Acc (\%)} & \textbf{F(3) (\%)} \\
\hline \hline
\multirow{3}{*}{0\%}  
& 1 & 3092 & 90.2 & 67.4 \\
& 2 & 2473 & 85.7 & 65.7 \\
& 3 & 698  & 84.4 & 66.5 \\
\hline
\multirow{3}{*}{5\%}  
& 1 & 2862 & 89.5 & 66.5 \\
& 2 & 1805 & 83.6 & 66.2 \\
& 3 & 645  & 80.5 & 65.7 \\
\hline
\multirow{3}{*}{10\%} 
& 1 & 1657 & 86.2 & 67.9 \\
& 2 & 1263 & 82.7 & 66.9 \\
& 3 & 454  & 74.7 & 65.1 \\
\hline
\end{tabular}
}
\vspace{-0.3cm}
\caption{Impact of hierarchical clustering thresholds and levels on the number of concept prototypes ($d_c$), classification accuracy, and interpretability (faithfulness F(3)) on the Cars dataset. Thresholds, expressed as a percentage of the maximum pairwise centroid distance.}
\label{tab:ward_cluster}
\end{table}

\subsubsection{(b) Effect of Modules: }
To assess the impact of each PCMNet component, we conduct an ablation study on the baseline PDiscoNet model without concept learning.  Additionally, the effect of using different types of concept vectors—prototypical and non-prototypical—is examined in Appendix B. Table~\ref{tab:modules_results} reports classification accuracy (Acc) and faithfulness (F(3)) on the Cars and Birds datasets. Adding the Marginal Clustering Center (MCC) loss improves accuracy (e.g., from 81.2\% to 87.8\% on Cars) and doubles faithfulness, highlighting the benefits of enforcing cluster-level consistency. Incorporating the Concept Mining Module (CMM) yields a major boost in faithfulness (from 4.9\% to 55.3\% on Birds), with minimal cost to accuracy. Combining both modules yields the best overall performance: 90.2\% accuracy and 67.4\% faithfulness on Cars, with only +0.4M parameters and +0.3G FLOPs.
Compared to recent interpretable baselines like PipPNet, ProtoPNet, ProtoViT, and MCPNet, PCMNet achieves a better balance between accuracy, faithfulness, and efficiency, especially outperforming ProtoViT in F(3) (+22.8\%) and requiring fewer FLOPs (17.4G vs. 24.8G). This confirms the complementary value of MCC and CMM in delivering interpretable and efficient visual recognition. The trade-off information is discussed in Appendix A.

\begin{table}[]
\resizebox{\linewidth}{!}{
\begin{tabular}{@{\extracolsep{0pt}}|l||ll|ll|ll|}
\hline
\multirow{2}{*}{\textbf{Settings}} & \multicolumn{2}{c|}{\textbf{Cars}} & \multicolumn{2}{c|}{\textbf{Birds}} & \multirow{2}{*}{\#\textbf{Params}} & \multirow{2}{*}{\textbf{FLOPs}} \\
                          & \textbf{Acc}        & \textbf{F(3)}       & \textbf{Acc}        & \textbf{F(3)}        &                          &                        \\ \hline \hline
Baseline                  & 81.2         & 5.4        & 82.3         & 4.9         & 27.1M                    & 17.1G                  \\
Baseline + MCC            & 87.8         & 9.2        & 84.9         & 8.9         & 27.2M                    & 17.1G                  \\
Baseline + CMM            & 85.8         & 58.6       & 82.5         & 55.3        & 27.4M                    & 17.4G                  \\
PCMNet (Full)       & 90.2         & \textbf{67.4}       & 85.1         & \textbf{64.9}        & 27.5M                    & \textbf{17.4G}                  \\ \hline
PiPNet       & 86.4         & 62.7      & 82.0         & 58.6        & 25.1M                    & 27.6G                  \\ 
ProtoPNet       & 84.5         & 38.6       & 81.4         & 31.7        & 28.9M                    & 18.4G                  \\ 
ProtoViT       & \textbf{91.8}         & 44.6      & \textbf{85.1}         & 36.7        & 22.1M                    & 24.8G                  \\ 
MCPNet       & 80.1         & 64.0      & 80.1         & 39.3        & 24.6M                    & \textbf{17.4G}                  \\ 
\hline

\end{tabular}
}
\vspace{-0.3cm}
\caption{Classification accuracy and faithfulness using different PCMNet modules.}
\label{tab:modules_results}
\end{table}

\subsection{Qualitative Results}
\begin{figure*}[!h]
    \centering
\includegraphics[trim={1.2cm 1.5cm 1cm 0.5cm},clip,width=.9\linewidth]{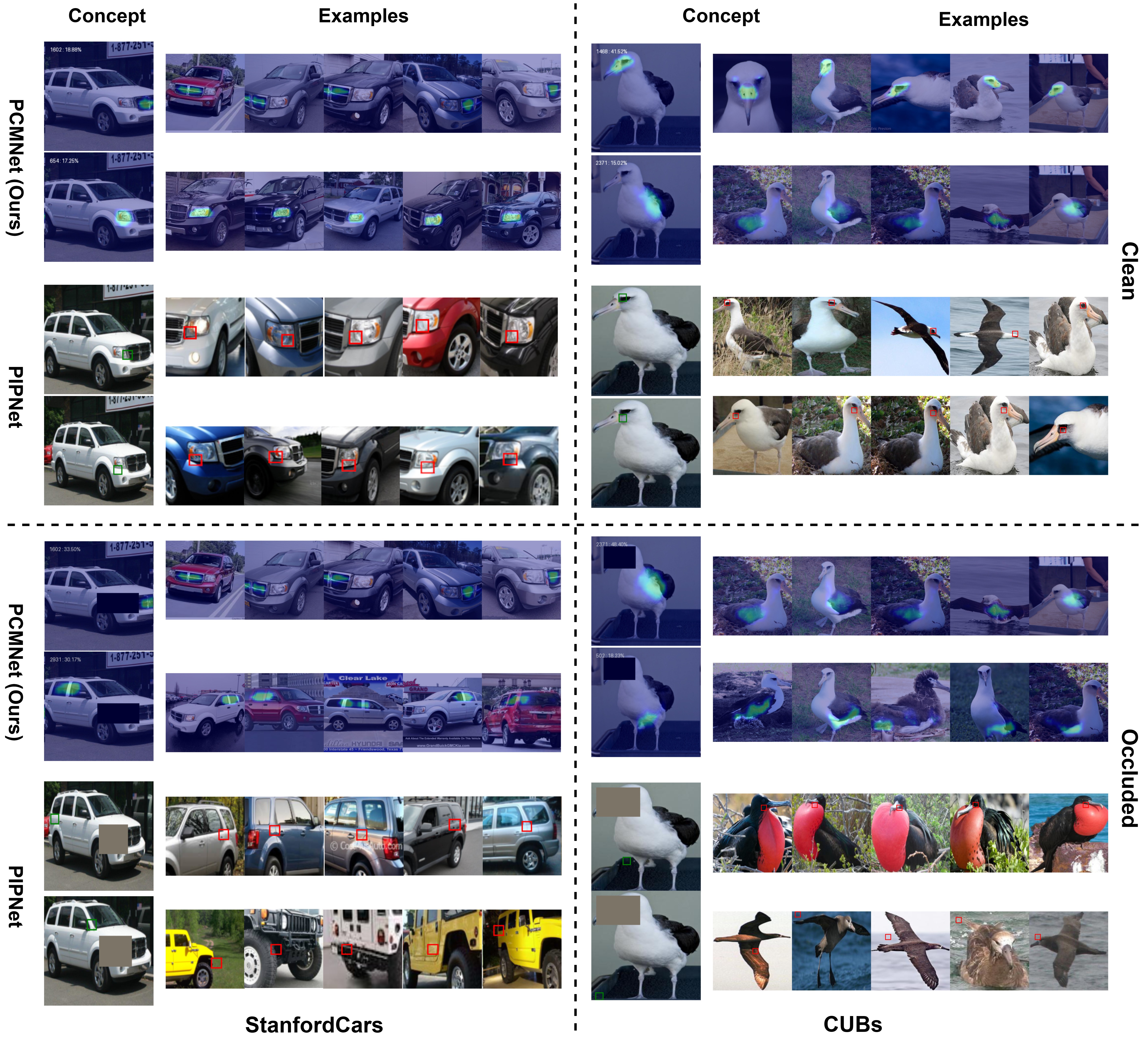}
\vspace{-0.1cm}
    \caption{visualizations of activated concept on test samples with matching training samples for Stanford Cars and CUB.}
    \vspace{-0.1cm}
    \label{fig:vis1}
\end{figure*}

\subsection{Robustness to Occlusion}
To evaluate PCMNet robustness under occlusion, we conducted controlled experiments by masking regions linked to the most activated concepts and re-evaluating the modified images with the model. This experiment aims to assess the stability of model predictions when key interpretable regions are removed. Specifically, we identify the center of the most activated concept—based on the model’s response to clean images—using the concept representation (patch for PIPNet and ProtoPNet, mask for PCMNet), and then occlude a rectangular region centered around this point. We consider three levels of occlusion, masking 10\%, 20\%, and 30\% of the image area. PCMNet's performance under these conditions is compared against two other interpretable baselines: ProtoPNet~\cite{ProtoPnet} and PIPNet~\cite{PIPNet}.


\begin{figure} [!ht]
  \centering
  \begin{subfigure}{0.45\linewidth}
    \includegraphics[trim={0.25cm 0.25cm 0.45cm 0.25cm},clip, width=\linewidth]{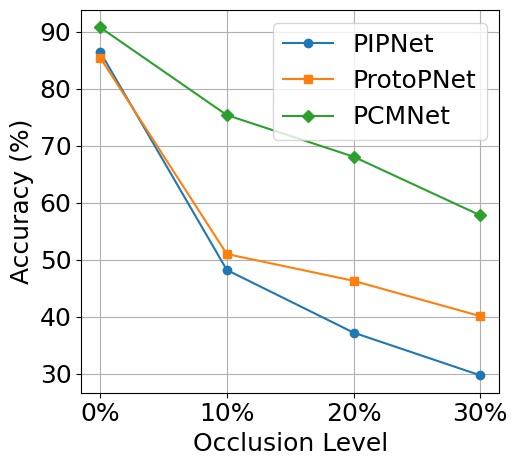}
    \vspace{-.45cm}
    \caption{Accuracy.}
  \end{subfigure}
  \hfill
  \begin{subfigure}{0.45\linewidth}
    \includegraphics[trim={0.25cm 0.25cm 0.45cm 0.25cm},clip,width=\linewidth]{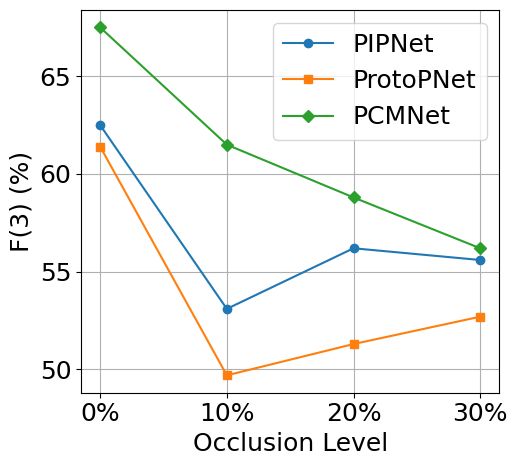}
    \vspace{-.45cm}
    \caption{Faithfulness F(3)}
  \end{subfigure}
    \vspace{-.25cm}
    \caption{Classification accuracy and faithfulness for a growing level of occlusion.}
    \label{fig:occlusion_results}
  \label{fig:lambda}
\end{figure}


Fig. \ref{fig:occlusion_results}.a shows that PCMNet degrades more gradually under occlusion compared to ProtoPNet and PIPNet. This robustness comes from PCMNet’s semantically rich and diverse part concepts, which capture complementary aspects of the object. Also, activated concepts from PCMNet preserve the faithfulness since they are extracted from not occluded regions as shown in Fig. \ref{fig:occlusion_results}.b and buttom of Fig. \ref{fig:vis1}.   
By describing class-discriminative features from multiple perspectives, PCMNet enables more reliable predictions. In contrast, patch-based models depend heavily on fixed, localized cues and thus exhibit sharper performance drops.

To further illustrate the interpretability and robustness of PCMNet, we compare the activated concepts of our method with PIPNet under clean and occluded conditions, as visualized in Fig. \ref{fig:vis1}. For each method, we select two activated concepts on test images and retrieve training set examples with the highest activation values for those concepts.
This comparison shows several key advantages of PCMNet:

\noindent \textit{Broader Concept Coverage:} PCMNet extracts more diverse and semantically rich concepts compared to PIPNet. It leverages attention-based masks instead of fixed patches to capture part-level semantics over a broader spatial context, leading to more varied and interpretable regions across different object parts and improving explanation quality.

\noindent \textit{Robustness to Occlusion:} As shown at the bottom of Fig. \ref{fig:vis1}, under occlusion, PIPNet often fails to highlight meaningful regions due to its reliance on limited regions. In contrast, PCMNet maintains interpretability by activating alternative, unoccluded parts that belong to the same or related concepts. This demonstrates the model’s ability to generalize concept reasoning even when certain visual cues are missing.
\begin{figure*}[!tbh]
    \centering
    \includegraphics[trim={0 0.22cm 0 0.15cm},clip,width=0.9\linewidth]{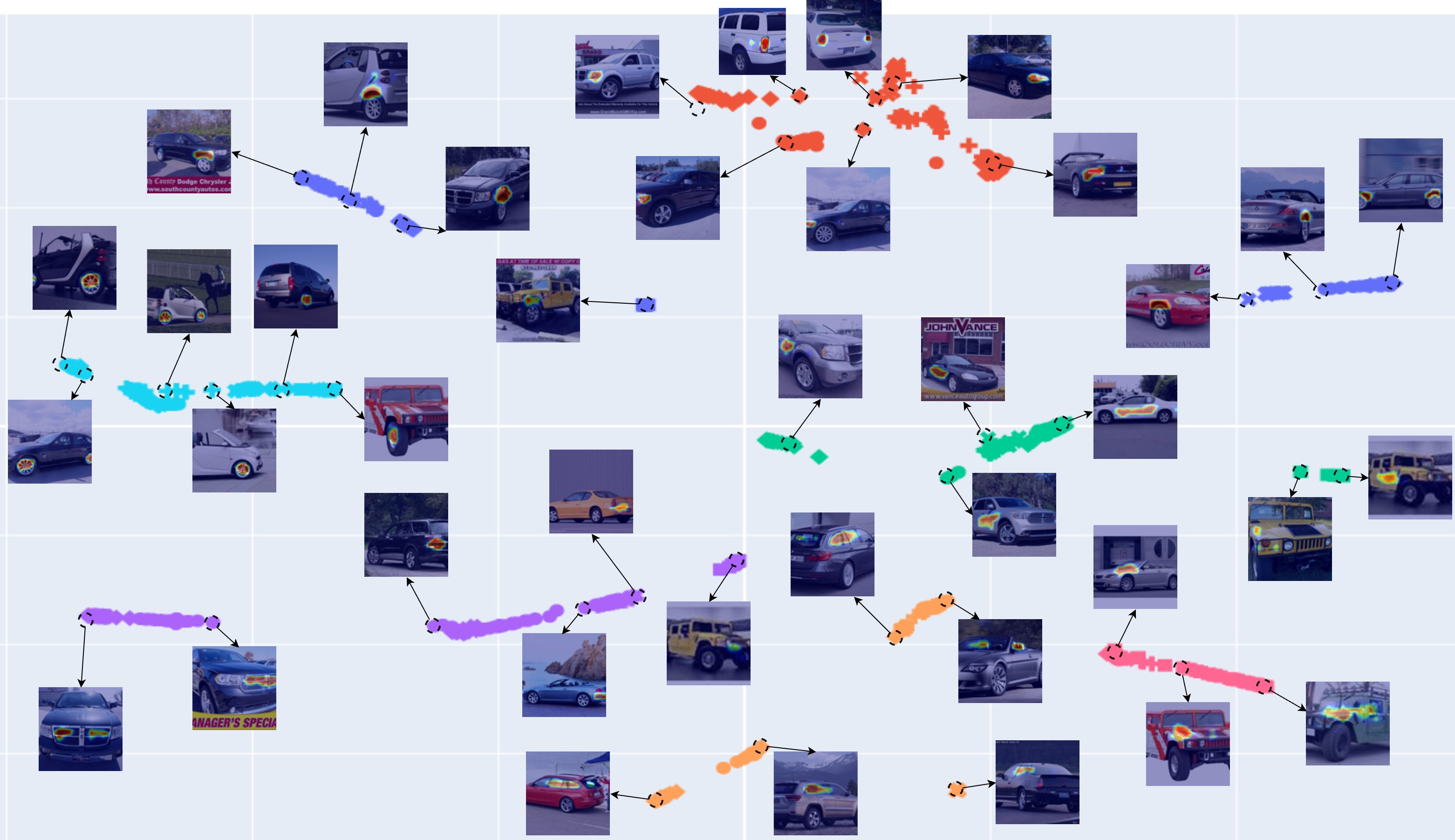}
    \vspace{-0.3cm}
   \caption{t-SNE visualization of concept-level part features extracted by PCMNet. Colors represent different prototypes, while shapes indicate object classes.}
    \label{fig:tsne}
\end{figure*}
To better understand the behavior of the learned prototypes in PCMNet, we visualize the extracted part-level features using t-SNE \cite{tsne}, as shown in Fig. ~\ref{fig:tsne}. Each point corresponds to a feature from a localized part, colors indicating the prototype (i.e. concept) and marker shapes denoting the object class. This visualization represents several insights:

\noindent \textit{Disentanglement of Prototypes:} Features assigned to different prototypes form well-separated clusters, indicating that PCMNet effectively disentangles part-based concepts. This shows that the Marginal Clustering Center loss and CMM promote diverse, semantically distinct parts, enhancing interpretability and occlusion robustness.
    
\noindent \textit{Multi-Class Concept Coverage:} Many prototype clusters contain the same semantic parts from multiple object classes. This indicates PCMNet learns class-agnostic concepts such as lights, wheels, or windows, which are shared across different categories. Such generalization is crucial for explainability, as it facilitates human-aligned, semantically meaningful concepts that transcend dataset labels.
    
\noindent \textit{Intra-Class Concept Diversity:} Interestingly, features from the same class are distributed across multiple prototype clusters. This highlights PCMNet's ability to capture \textit{intra-class variation} through multiple part-based concepts. For instance, different views or structural variations of a car class may be assigned to distinct concepts (Backlight vs headlight), further enhancing the model’s fine-grained interpretability based on human perception and explanation.



\section{Conclusion}
We presented PCMNet, a novel framework designed to bridge the gap between accuracy and interpretability in deep learning models. By mining part-prototypes from dynamic image regions, PCMNet improves upon existing patch-based prototype models by ensuring semantic coherence across instances. Our results show that PCMNet enhances faithfulness, concept stability, and explainability, outperforming existing interpretable models. Moreover, our occlusion-based robustness analysis confirmed that PCMNet remains stable under missing or distorted input regions, demonstrating its reliability in real-world scenarios. Unlike conventional post-hoc explainability methods, PCMNet allows for more human-aligned, concept-based reasoning, making deep learning decisions more transparent and interpretable.
\label{sec:conclusion}

\paragraph{Acknowledgment:} This research was supported by the Natural Sciences and Engineering Research Council of Canada (NSERC), and the Digital Research Alliance of Canada.


\bibliography{aaai2026}

\onecolumn
\clearpage 
\newpage
\appendix
\section{Supplementary: Additional Experiment and Results}
\label{sec:appendix_section}

\subsection{Appendix A: Additional Implementation Details}

In our model, $F^i = \mathcal{F}(x^i)$ denotes the spatial feature map extracted by the backbone network $\mathcal{F}$, with shape $\mathbb{R}^{D \times w \times h}$, where $D$ is the backbone feature dimension, and $w \times h$ is the spatial resolution.
The part-prototype module $\mathcal{P}$ produces a soft assignment mask $M^i = \mathcal{P}(F^i) \in \mathbb{R}^{(K+1) \times w \times h}$, where each pixel location is softly assigned to one of the $K$ part and one background channels. Specifically, for a given 2D location $a = (a^x, a^y)$, the assignment scores across all parts are:
\begin{equation}
    M^i(a) = M^i[:, a^x, a^y] \in \mathbb{R}^{K+1}.
\end{equation} 
while the full mask for part $p$ is given by, 
\begin{equation} 
    M^i_p = M^i[p, :, :] \in \mathbb{R}^{w \times h}.
\end{equation} 

We used Resnet50 as the backbone, which extracts 2048 features. We project these features into 512 features before global pooling, i.e., $d_f=512$. Based on our experiment and set of hyper-parameters, for the CUBs dataset, the $d_c=2954$ and Stanford Cars $d_c=3092$ are computed in stage 2. As the DBSCAN is used for clustering inside of each class, for each part and class, the number of clusters is not fixed. 

\subsubsection{Learning Curvature}
Fig.~\ref{fig:learning_curve} illustrates the training loss and validation accuracy of PCMNet over 40 epochs. In Step 1, only the backbone and part-discovery modules are trained using the part discrimination loss ($\mathcal{L}\text{part}$). Once this loss converges, Step 2 and 3 begin, during which concept centroids are extracted and the clustering loss ($\mathcal{L}\text{cl}$) is introduced, respectively. This causes a temporary spike in the loss, followed by a gradual decrease as the model adapts to reasoning over learned concepts. To keep the prototypes dynamic and data-aligned, centroid extraction and Step 2 are repeated every 5 epochs.
\begin{figure}[!h]
    \centering
    \includegraphics[width=0.6\linewidth]{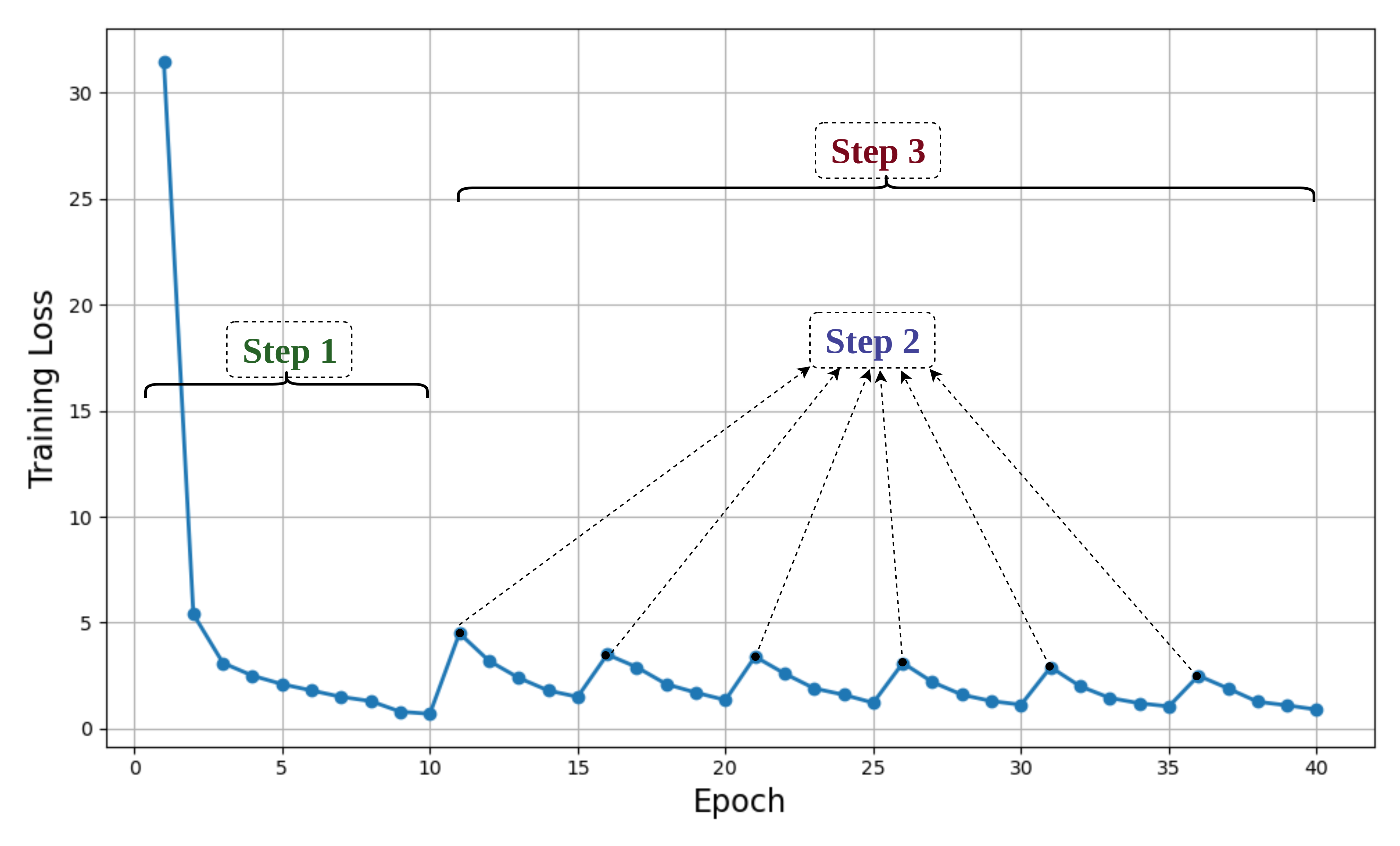}
    \caption{Training loss over 40 epochs showing three-stage learning dynamics.}
    \label{fig:learning_curve}
\end{figure}

\subsubsection{Additional details of Baseline loss}
The $\mathcal{L}_\text{bl}$\cite{van2023pdisconet} loss also is 
\begin{equation}
    \mathcal{L}_\text{bl} =   \mathcal{L}_\text{cond} +  \mathcal{L}_\text{equi} + \mathcal{L}_\text{pres},
\end{equation}
where the objectives for each loss are:
\begin{itemize}
    \item  $\mathcal{L}_{cond}$: Each detected part should consist of a condensed and contiguous image region.
    \item  $\mathcal{L}_{equi}$:  encourages the equivariance of the attention maps in for input images and their augmented version (translation, rotation or scaling).
    \item $\mathcal{L}_{pres}$: All parts should be present at least in some of the images in the batch to avoid finding non-relevant parts. 
\end{itemize}

\subsection{Appendix B: Additional Ablation Studies }

\subsubsection{Number of Parts ($K$)}
We assess the impact of the number of parts on classification accuracy, faithfulness, and size of concept prototypes in Table~\ref{tab:part}. As the model extracts more prototypical concepts, its decisions become more aligned with the underlying features, enhancing interpretability. Interestingly, when the number of prototypes is smaller, the faithfulness scores for the top concepts tend to be higher. This is because the model relies on fewer, more compact concepts ($d_c$), concentrating its decisions on a smaller number of more influential ones.

\begin{table}[!b]
\small
\centering
{
\begin{tabular}{|l||c|c|c||c|c|c|}
\hline
\multirow{2}{*}{\textbf{K}} & \multicolumn{3}{c||}{Cars} &  \multicolumn{3}{c|}{Birds} \\ \cline{2-7}
 & \textbf{C Acc} & \textbf{F(1)-F(5)} & $d_c$ & \textbf{C Acc} & \textbf{F(1)-F(5)} & $d_c$ \\ \hline \hline
0   & 84.1    & 1.54 - 6.61 & - & 83.2 &  1.21-5.56       & - \\ \hline
4           & 83.5     & \textbf{79.96} - 91.01 & 1703 & 83.5 & \textbf{71.42}-85.64   & 1786    \\ 
5           & 84.1     & 70.11 – 91.33 & 1979  & 84.9 &   66.7 - 85.9 & 2018   \\ 
6           & 85.2     & 66.02 – 90.65 & 2345 & 85.7 &  59.9 - 86.7  & 2597  \\ 
7           & 86.7     & 62.47 – 91.25 & 2767 & \textbf{86.0}  &  55.7 - \textbf{87.3} & 2954  \\ 
8           & \textbf{87.5}     & 59.33 – \textbf{91.73} & 3092  & 85.1 & 53.0 - 86.4 & 3242   \\ 
9           & 86.4     &  55.10 – 91.52 & 3277 & 84.4 &  52.5-86.1  & 3470     \\ \hline
\end{tabular}
}
\vspace{-0.16cm}
  \caption{Impact on performance of the number of prototypes.}
  \label{tab:part}
\end{table}

\subsubsection{Effect of Hyper-parameters.}
We investigate the sensitivity of our model to the hyper-parameters \( \alpha \) and \( \beta \), which control the strength of the concept consistency and interpretability regularization terms, respectively. As shown in Fig.~\ref{fig:alpha_beta}, our model remains robust across a wide range of values. Specifically, the classification accuracy (C Acc) and faithfulness score \textbf{F(\textit{n})} exhibit stable trends, with peak performance achieved around \( \alpha = 1.5 \) and \( \beta = 2.0 \). This indicates that PCMNet effectively balances accuracy and interpretability without requiring extensive hyper-parameter tuning. The consistency of the scores across different settings highlights the reliability of the proposed regularization components.

\begin{figure}[!h]
    \centering
    \includegraphics[width=0.8\linewidth]{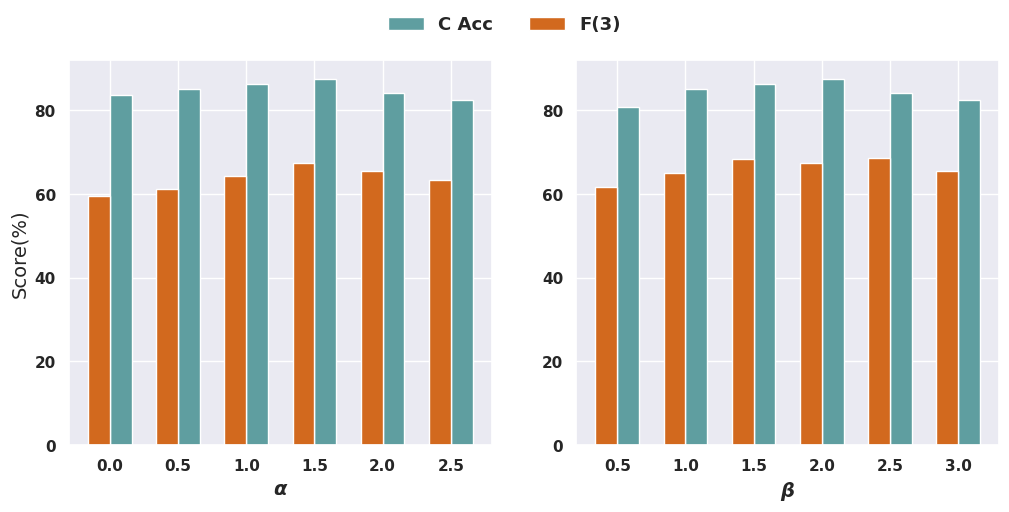}
    \caption{Impact of hyper-parameters \( \alpha \) and \( \beta \) on classification accuracy (C Acc) and faithfulness score \textbf{F(\textit{3})}. Our method maintains strong performance across a wide range of values, showing robustness and stability in both predictive accuracy and interpretability.}
    \label{fig:alpha_beta}
\end{figure}

Also, in Fig. \ref{fig:lambda}, we evaluate the impact of hyperparameters in Eq.\ref{eq:cen} on classification accuracy.
At first, we set $m_2=1$ and measure the effect of $m_1$ by gradually increasing the values. Once it is found, we search for $m_2$. The best performance is achieved when $m_1$ is set to 0.3 and $m_2$ to 1.5, respectively. The upward trend of the bars demonstrates the effectiveness of each loss. 
\begin{figure} [!ht]
  \centering
  \begin{subfigure}{0.4\linewidth}
    \includegraphics[width=\linewidth]{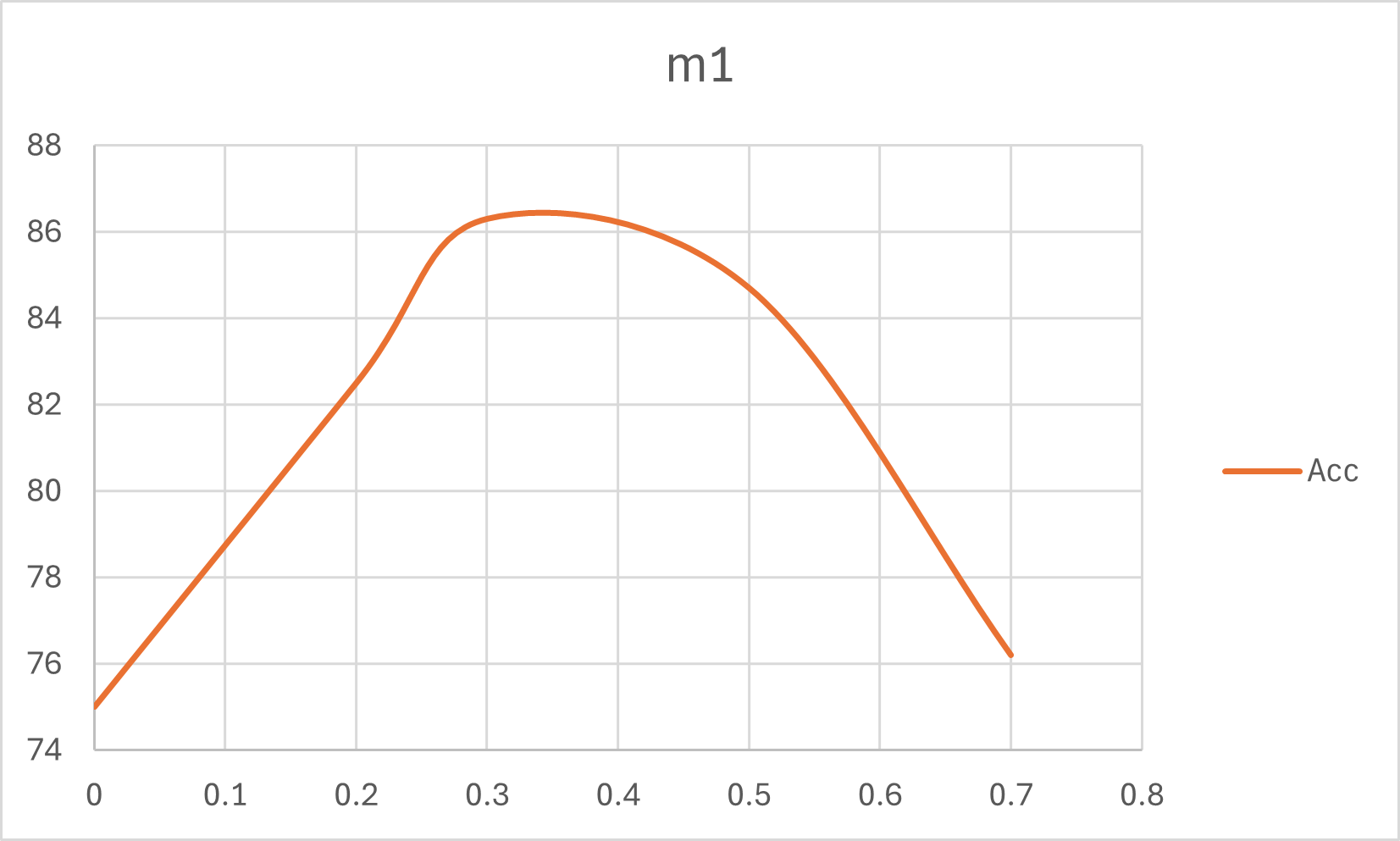}
    \caption{$m1$.}
  \end{subfigure}
  \hfill
  \begin{subfigure}{0.4\linewidth}
    \includegraphics[width=\linewidth]{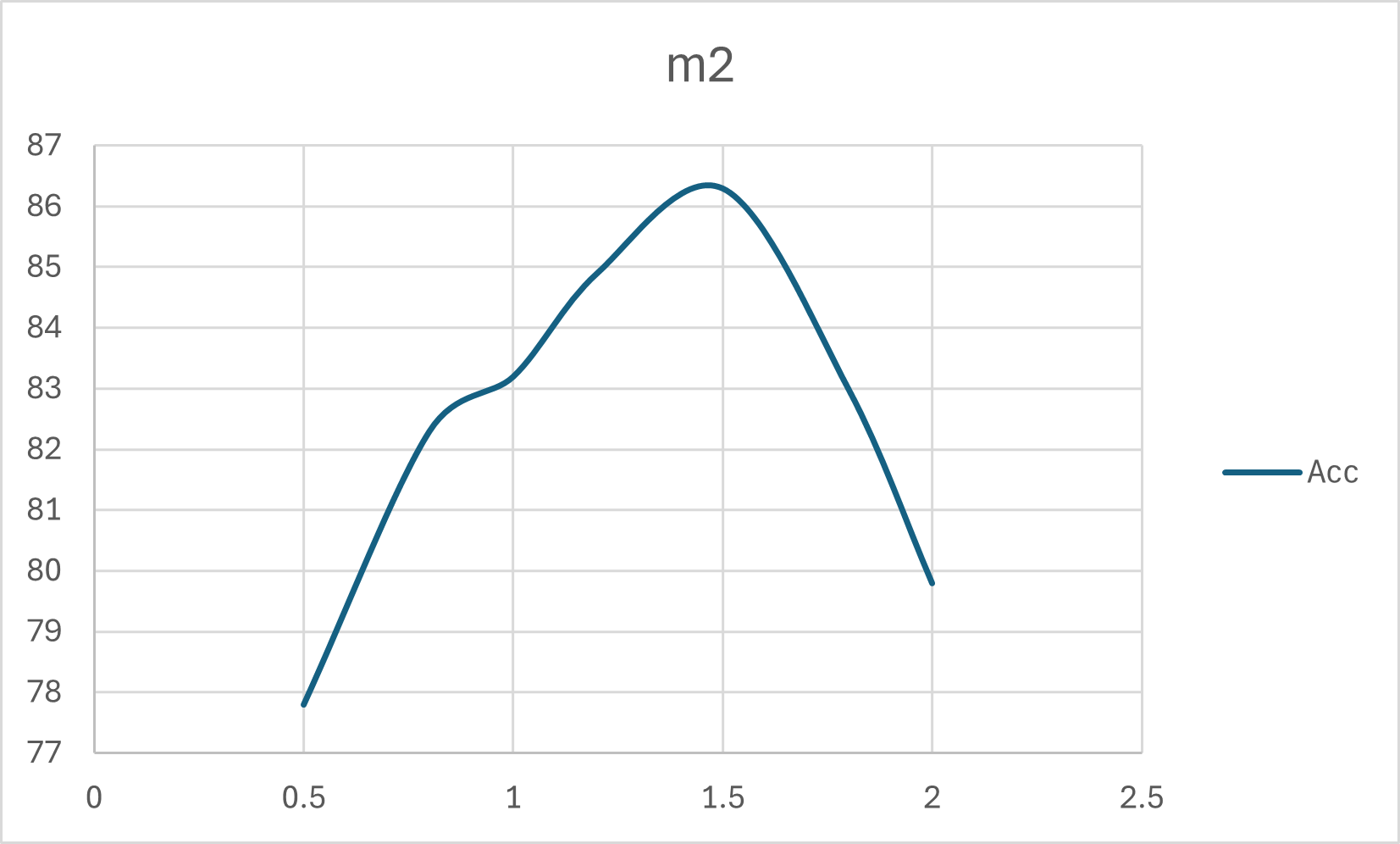}
    \caption{$m2$}
  \end{subfigure}
  \caption{ The model's classification accuracy in different values of $m1$ and $m2$. }
  \label{fig:lambda}
\end{figure}

\subsubsection{Effect of Non-Prototypical Concepts }
To evaluate the impact of different concept vector types on model performance and interpretability, we conduct an ablation study using prototypical (\(\mathbf{z^i}\)) and non-prototypical (\(\mathbf{g^i}\)) concept vectors. As shown in Table~\ref{tab:g_p}, using only non-prototypical vectors significantly reduces both classification accuracy and faithfulness scores. In contrast, employing only prototypical vectors retains higher classification performance and strong concept-based faithfulness. The full PCMNet, which combines both concept types, achieves the best results across both datasets, suggesting that the interplay between prototypical and non-prototypical representations enhances both prediction accuracy and interpretability.
In other words, the vector $\mathbf{g}^i$ is used to capture features that are not assigned to any prototypical part. Importantly, these features do not necessarily correspond to background pixels. Instead, they include informative regions of the object that are not consistently detected as one of the predefined parts by the part-detection module.
As illustrated in Fig. \ref{fig:vis3}, these non-prototypical concepts often correspond to distinctive yet less frequent features that are still valuable for decision-making. For example:

\begin{itemize}
    \item In the Cars dataset, the ``cargo'' region is activated as a non-prototypical concept. Though it appears in a small subset of vehicles only, that provides strong discriminative power.
    \item In the CUB dataset, the ``webbed feet'' of certain birds (e.g., Albatross species) are highlighted. These regions are highly specific and explainable, yet do not form a consistent enough prototype across classes to be captured in the prototypical set.
\end{itemize}

\begin{table}[!h]
\small
\centering
{
\begin{tabular}{|l||c|c||c|c|}
\hline
\multirow{2}{*}{\textbf{CAVs}} & \multicolumn{2}{c||}{Cars} &  \multicolumn{2}{c|}{Birds} \\ \cline{2-5}
 & \textbf{C Acc} & \textbf{F(1)-F(5)} & \textbf{C Acc} & \textbf{F(1)-F(5)} \\ \hline
Baseline          & 84.1    & 1.54 - 6.61  & 83.2 &  1.21-5.56       \\ 
PCMNet + Prototypical           & 86.2     & 61.8 – 92.63 & 84.5  &  56.8 - 87.4 \\
PCMNet + Non-Prototypical           & 79.4     & 42.80 – 75.73  & 78.15  &  35.9 - 67.7 \\

PCMNet (Full)           & \textbf{87.5}     & \textbf{59.33} – \textbf{91.73}  & \textbf{86.0}  &  \textbf{55.7} - \textbf{87.3} \\\hline

\end{tabular}
}
  \caption{Impact of prototypical (\(\mathbf{z^i}\)) and non-prototypical (\(\mathbf{g^i}\)) concept vectors on PCMNet’s classification accuracy (C Acc) and faithfulness scores (F(1)-F(5)) on Cars and Birds datasets. The full model achieves superior performance, highlighting the benefit of combining both concept types.}

  \label{tab:g_p}
\end{table}

\subsection{Comparison of Performance, Interpretability, and Efficiency}
To better visualize the trade-offs among classification performance, interpretability, and computational efficiency, we present a summary plot comparing PCMNet to recent interpretable AI baselines (Fig.~\ref{fig:method_tradeoff_plot}). Each method is represented using a unique shape, and the color intensity of each point reflects its FLOPs (darker points indicate higher computational cost). The x-axis corresponds to classification accuracy on the Stanford Cars dataset, while the y-axis shows faithfulness under occlusion (F(3)), one of our main interpretability metrics.

\begin{figure}[!h]
    \centering
    \includegraphics[width=0.8\linewidth]{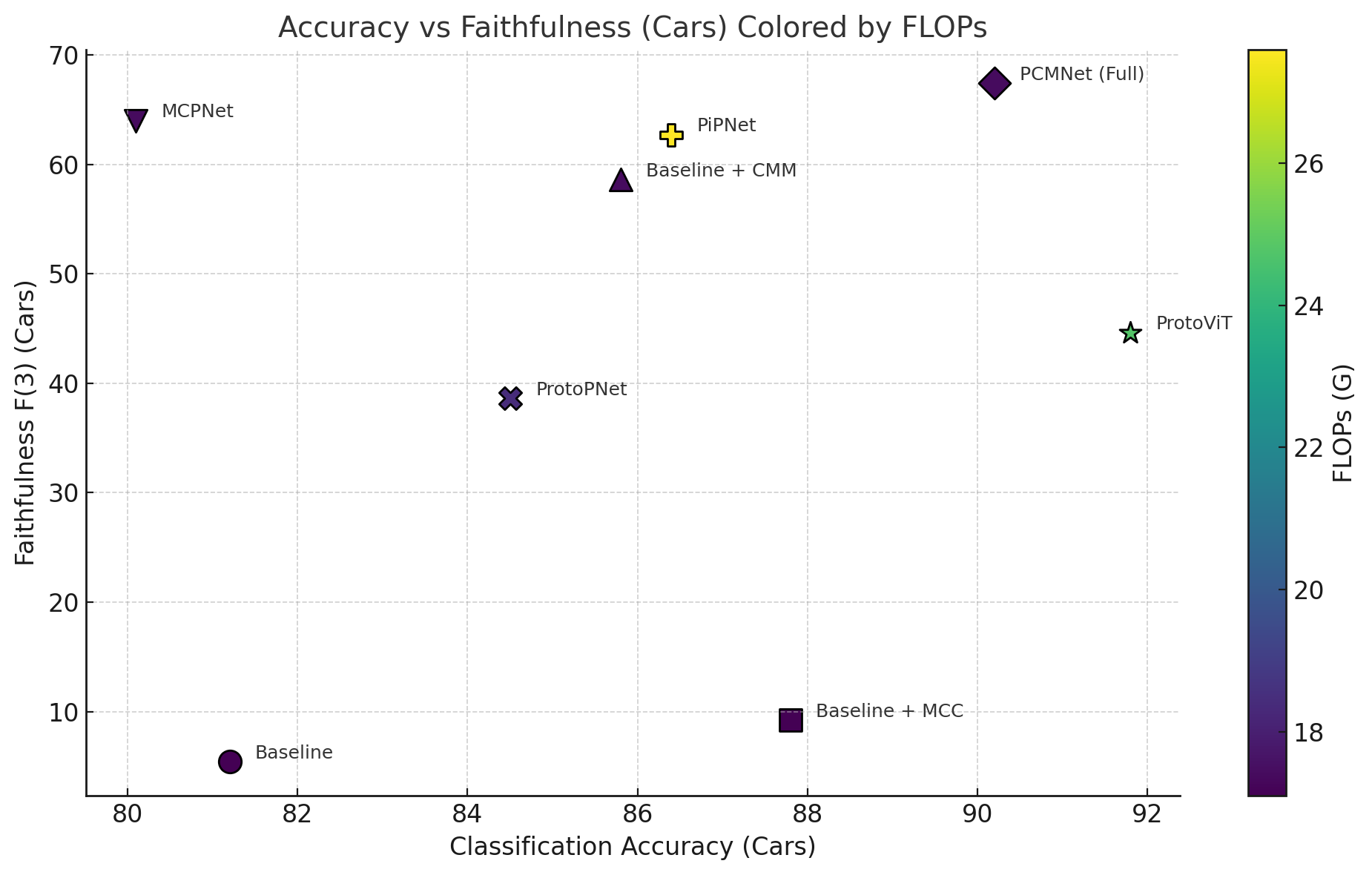}
    \caption{Trade-off comparison between accuracy, faithfulness (F(3)), and efficiency (FLOPs) across methods on the Stanford Cars dataset.
Each method is represented by a distinct marker shape, and the color intensity encodes its computational cost (darker = less FLOPs). PCMNet achieves the best balance of high accuracy, top interpretability, and low computational cost, demonstrating its practical advantage over state-of-the-art interpretable models.}
    \label{fig:method_tradeoff_plot}
\end{figure}

PCMNet (full) achieves an excellent balance: it attains competitive accuracy (90.2\%) and the highest interpretability (F(3) = 67.4) among all methods, while maintaining one of the lowest FLOPs (17.4G). In contrast, ProtoViT achieves slightly higher accuracy (91.8\%) but suffers from lower interpretability (F(3) = 44.6) and significantly higher FLOPs (24.8G). ProtoPNet and PiPNet either underperform in accuracy or require substantially more computation.
This visualization supports our claim that PCMNet is highly practical: it delivers state-of-the-art interpretability without sacrificing classification performance or efficiency.

\subsection{Appendix C: Additional Qualitative Results }
More results of the PCMNet prototypical concept are shown in Figs. \ref{fig:vis11} and \ref{fig:vis2}. Fig \ref{fig:vis3} illustrates examples of non-prototypical concepts discovered by our model for the Cars (left) and CUBs (right) datasets. For each dataset, the concept location on a test image is highlighted with a green bounding box, while corresponding activations in training examples are shown on the right using red bounding boxes. These concepts provide complementary cues to prototypical ones. For example, in the first row of Cars, the "cargo" attributes are not well represented in prototypical concepts since they appear in a small subset of vehicles, yet they are highly discriminative for decision-making. Similarly, the last row in CUBs shows the "webbed feet" concept, which is especially relevant and explainable for Albatross species, while foot prototypes are not sufficiently distinctive across the dataset. These findings highlight the value of incorporating non-prototypical concepts in improving both interpretability and robustness of the model.

\begin{figure*}[!ht]
    \centering
    \includegraphics[width=\linewidth]{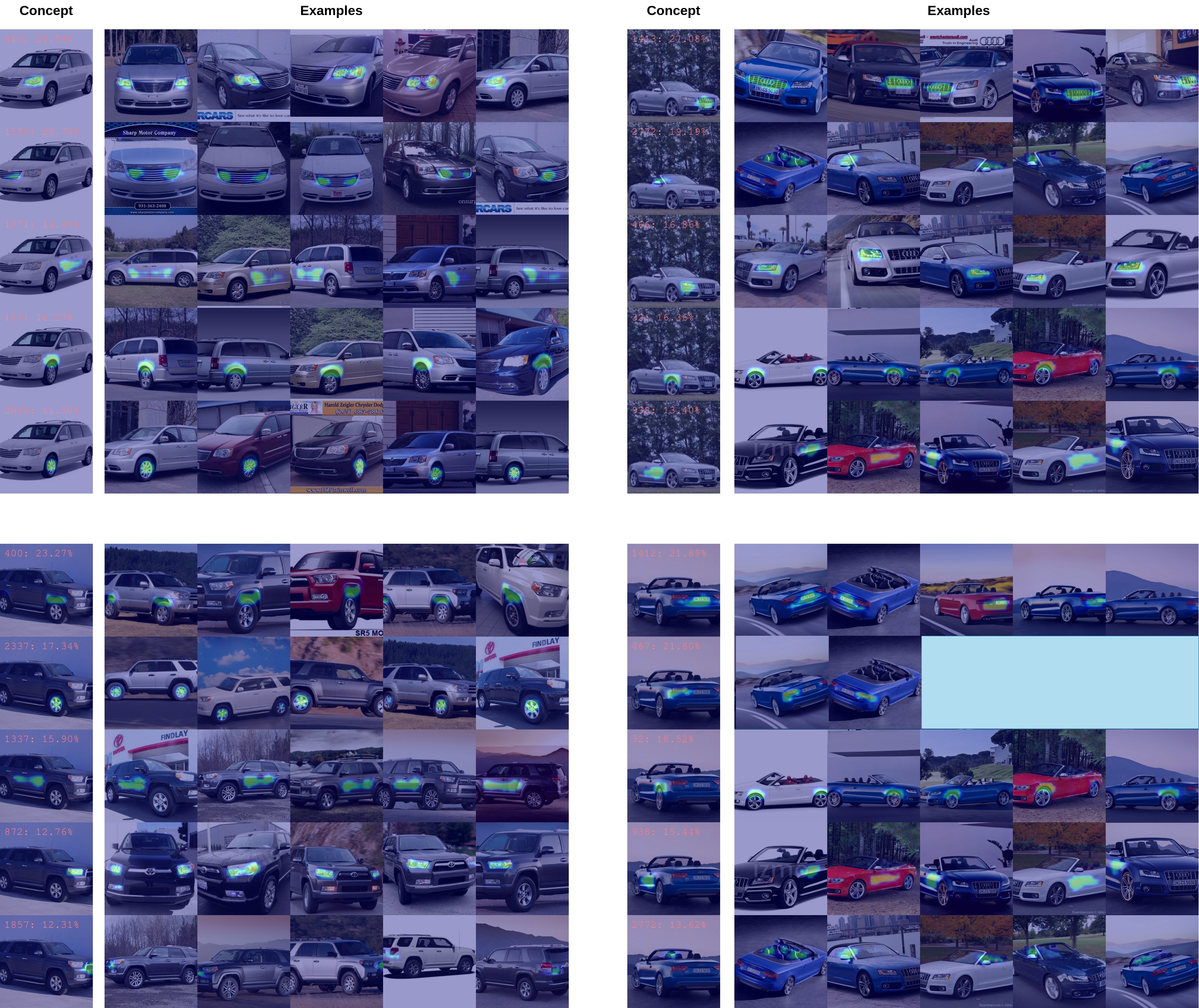}
    \caption{Visualization of activated concepts in test data and showing some examples in the training set for StanfordCars datasets. Each concept is shown with an index and percentage of explanation for the model decision.}
    \label{fig:vis11}
\end{figure*}

\begin{figure*}[!hb]
    \centering
    \includegraphics[width=\linewidth]{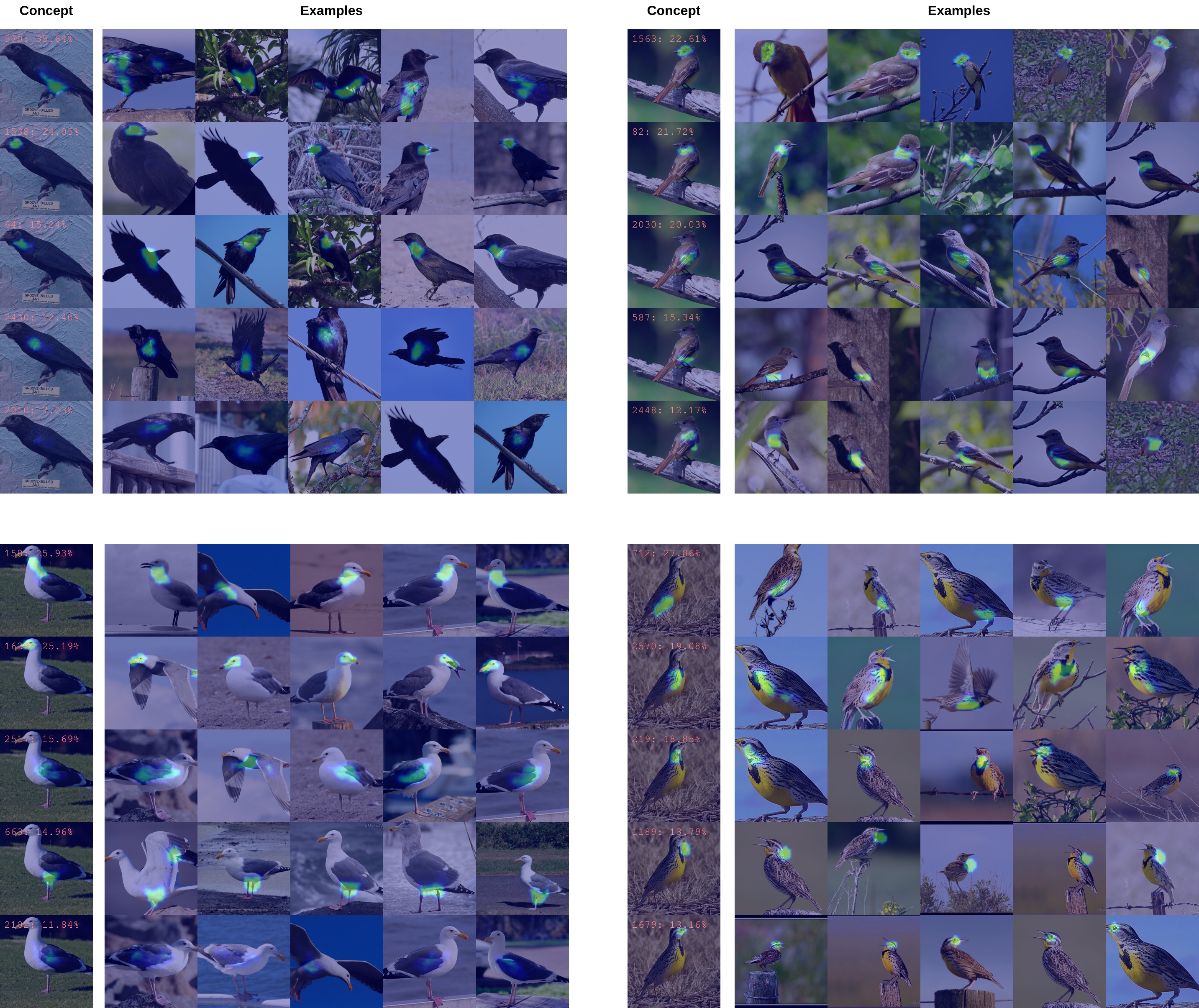}
    \caption{Visualization of activated concepts in test data and showing some examples in the training set for CUB-200 Birds datasets. Each concept is shown with an index and percentage of explanation for the model decision.}
    \label{fig:vis2}
\end{figure*}

\begin{figure*}[!b]
    \centering
    \includegraphics[width=\linewidth]{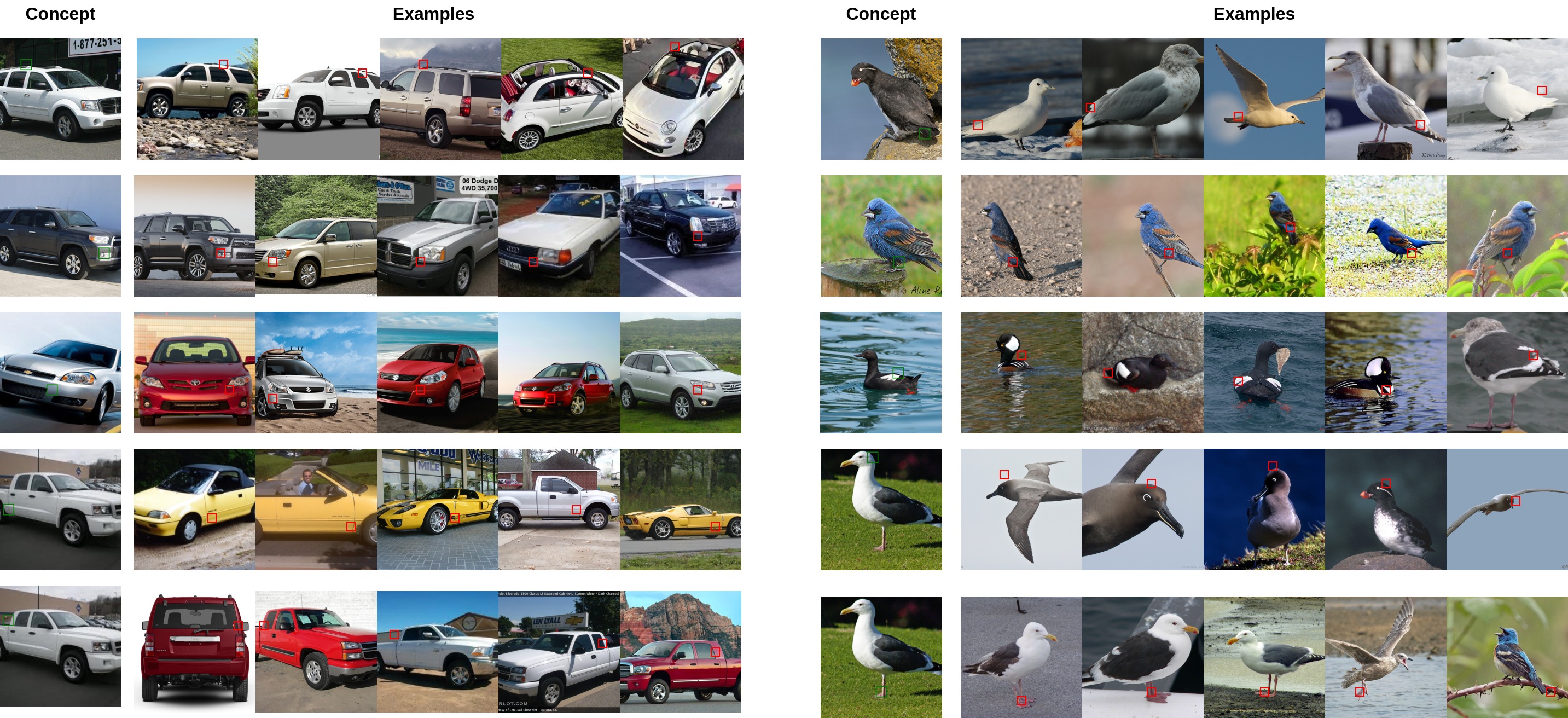}
    \caption{Visualization of non-prototypical activated concepts in test data and showing some examples in the training set for (left) Stanford Cars and (right) CUB-200 Birds datasets. Similar to ProtoPNet and PIPNet, we select the patch for the region of the activated concept.}
    \label{fig:vis3}
\end{figure*}


\end{document}